%% file: main.tex
\documentclass{article}

\usepackage{microtype}
\usepackage{graphicx}
\usepackage{booktabs} 
\usepackage[normalem]{ulem}
\usepackage[export]{adjustbox}
\usepackage{multirow}
\usepackage{stfloats}
\usepackage{babel,blindtext}
\usepackage{pgfplots}
\usepackage{subcaption,graphicx}
\usepackage{lipsum}
\usepackage{caption}
\usepackage{rotating}
\usepackage{array}
\usepackage{enumitem}

\usepackage{hyperref}

\usepackage[accepted]{icml2023}

\usepackage{amsmath}
\usepackage{amssymb}
\usepackage{mathtools}
\usepackage{amsthm}
\usepackage{bm}

\usepackage[capitalize,noabbrev]{cleveref}

\theoremstyle{plain}

\theoremstyle{definition}

\theoremstyle{remark}

\usepackage[textsize=tiny]{todonotes}

\icmltitlerunning{DDPMs for Robust Image Super-Resolution in the Wild\hfill\thepage}

\begin{document}

\twocolumn[
\icmltitle{Denoising Diffusion Probabilistic Models for Robust Image \\ Super-Resolution in the Wild}




\begin{icmlauthorlist}
\icmlauthor{Hshmat Sahak}{brain,intern}
\icmlauthor{Daniel Watson}{brain}
\icmlauthor{Chitwan Saharia}{brain}
\icmlauthor{David Fleet}{brain}
\end{icmlauthorlist}

\icmlaffiliation{brain}{Google Research, Brain Team.}
\icmlaffiliation{intern}{Work done as a student researcher}

\icmlcorrespondingauthor{Hshmat Sahak}{hshmat.sahak@mail.utoronto.ca}

\vskip 0.3in
]



\printAffiliationsAndNotice{}  

\begin{abstract}
    Diffusion models have shown promising results on single-image super-resolution and other image-to-image translation tasks.
    Despite this success, they have not outperformed state-of-the-art GAN models on the more challenging {\em blind super-resolution} task, where the input images are out of distribution, with unknown degradations. 
    This paper introduces SR3+, a diffusion-based model for blind super-resolution, establishing a new state-of-the-art.
    To this end, we advocate self-supervised training with a combination of composite, parameterized degradations for self-supervised training, and noise-conditioing augmentation during training and testing.
    With these innovations, a large-scale convolutional architecture, and large-scale datasets, SR3+ greatly outperforms SR3.
    It outperforms Real-ESRGAN when trained on the same data, with a DRealSR FID score of 36.82 vs.\ 37.22, which further improves to FID  of 32.37 with larger models, and further still with larger training sets.
\end{abstract}

\input{images/representative_samples.tex}

\input{introduction.tex}

\input{background.tex}

\input{relatedwork.tex}

\input{methodology.tex}

\input{experiments.tex}
\input{conclusion.tex}
\bibliography{citations}
\bibliographystyle{icml2023}

\newpage
\appendix
\onecolumn
\input{appendix.tex}


\end{document}

%% file: images/representative_samples.tex
\begin{figure}
\vspace*{-0.1cm}
\hspace*{-0.2cm}
\resizebox{0.952\columnwidth}{!}{%
\begin{tabular}{m{0.28cm} m{0.35\columnwidth} m{0.35\columnwidth}}

  \rotatebox{90}{{\footnotesize Input image}}
& \includegraphics[height=1.25in]{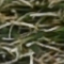}
& \includegraphics[height=1.25in]{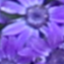}
\\
  \rotatebox{90}{{\footnotesize SR3}}
& \includegraphics[height=1.25in]{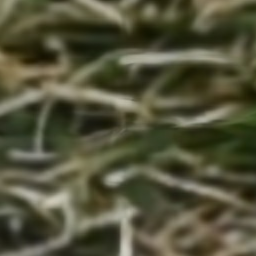}
& \includegraphics[height=1.25in]{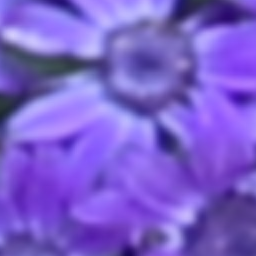}
\\
  \rotatebox{90}{{\footnotesize  Real-ESRGAN}}
& \includegraphics[height=1.25in]{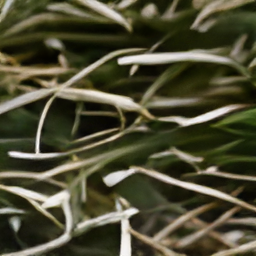}
& \includegraphics[height=1.25in]{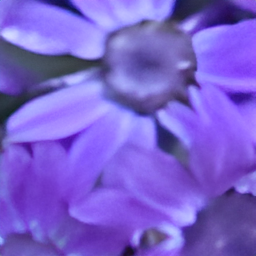}
\\
  \rotatebox{90}{{\footnotesize \textbf{SR3+ (Ours)}}}
& \includegraphics[height=1.25in]{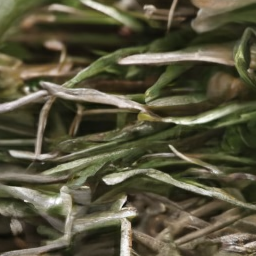}
& \includegraphics[height=1.25in]{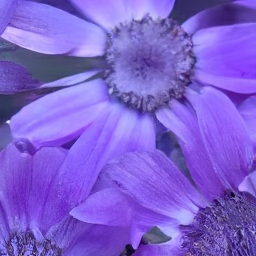}
\\
  \rotatebox{90}{{\footnotesize Ground Truth}}
& \includegraphics[height=1.25in]{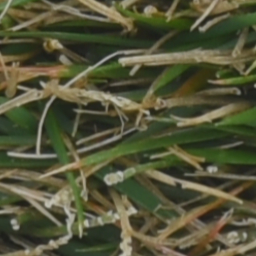}
& \includegraphics[height=1.25in]{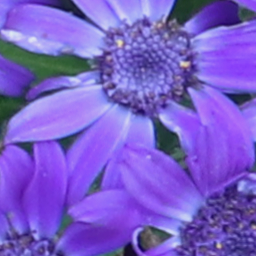}
\\
\end{tabular}%
}
\vspace*{-0.2cm}
\caption{Blind super-resolution test results ($64\!\times\!64 \rightarrow 256\!\times\! 256$) for SR3+, SR3 and Real-ESRGAN.}
\label{fig:representative-samples}
\vspace*{-0.25cm}
\end{figure}

%% file: introduction.tex
\section{Introduction}
\label{section:introduction}

Diffusion models \cite{sohl2015deep,song2019generative,ho2020denoising,song2020score} have quickly emerged as a powerful class of generative models, advancing the state-of-the-art for both text-to-image synthesis and image-to-image translation tasks 
\cite{dhariwal2021diffusion,rombach2022high,saharia2022palette,saharia2022image,li2022srdiff}.
For single image super-resolution \citet{saharia2022image}, showed strong performance
with self-supervised diffusion models, leveraging  their ability to capture complex multi-modal distributions, typical of 
super-resolution tasks with large magnification factors.
Although impressive, SR3 falls short on out-of-distribution (OOD) data, i.e., images in the wild with unknown degradations. Hence
GANs remain the method of choice for \textit{blind super-resolution} \cite{wang2021real}.

This paper introduces SR3+, a new diffusion-based super-resolution model that is both flexible and robust, achieving state-of-the-art results on OOD data (Fig.\ \ref{fig:representative-samples}).
To this end, SR3+ combines a
simple convolutional architecture and a novel training process with two key innovations.
Inspired by \citet{wang2021real} we use parameterized degradations in the data augmenation training pipeline, with significantly more complex corruptions in the generation of low-resolution (LR) training inputs compared to those of \cite{saharia2022image}.
We combine these degradations with \textit{noise conditioning augmentation}, first used to improve robustness in cascaded diffusion models \citet{ho2022cascaded}.
We find that noise conditioning augmentation is also effective at test time for zero-shot application.
SR3+ outperforms both SR3 and Real-ESRGAN on FID-10K 
when trained on the same data, with a similar sized model, and applied in zero-shot testing on both the RealSR \cite{cai2019toward} and DRealSR \cite{wei2020component} datasets.
We also show further improvement simply by increasing model capacity and training set size.

Our main contributions are as follows: 
\vspace*{-0.33cm}
\begin{enumerate}[leftmargin=0.5cm]
    \itemsep 0.015cm
    \item We introduce SR3+, a diffusion model for blind image super-resolution, outperforming SR3 and the previous SOTA on zero-shot RealSR and DRealSR benchmarks, across different model and training set sizes.
    \item Through a careful ablation study, we demonstrate the complementary benefits of parametric degradations and noise conditioning augmentation techniques (with the latter also used at test time).
    \item We demonstrate significant improvements in SR3+ performance with increased model size, and with larger datasets (with up to 61M images in our experiments).
\end{enumerate}

%% file: background.tex
\section{Background on Diffusion Models}
\label{section:background}
Generative diffusion models are trained to learn a data distribution in a way that allows computing samples from the model itself. This is achieved by first training a \textit{denoising} model.
In practice, given a (possibly conditional) data distribution $q(\bm{x}|\bm{c})$, one constructs a Gaussian \textit{forward process}
\begin{equation}
\label{eqn:forward}
q(\bm{z}_t | \bm{x}, \bm{c}) = \mathcal{N}(\bm{z}_t; \sqrt{\alpha_t} \bm{x}, (1-\alpha_t) \bm{I})
\end{equation}
where $\alpha_t$ is a monotonically decreasing function over $t \in [0, 1]$, usually pinned to $\alpha_0 \approx 1$ and $\alpha_1 \approx 0$. At each training step, given a random $t \sim \mathrm{Uniform}(0, 1)$, the neural network $\bm{x}_\theta(\bm{z}_t, t, \bm{c})$ must learn to map the noisy signal $\bm{z}_t$ to the original (noiseless) $\bm{x}$. \citet{ho2020denoising} showed that a loss function that works well in practice is a reweighted evidence lower bound \citep{kingma2013auto}:
\begin{equation}
    L(\theta) = \mathbb{E}_{\bm{x},t,\bm{\epsilon}} \| \bm{\epsilon}_\theta(\bm{z}_t, t, \bm{c}) - \bm{\epsilon} \|^2
\end{equation}

where the neural network learns to infer the additive noise $\epsilon$, as opposed to the image itself.
Recovering the image is then trivial, since we can use the reparametrization trick \citep{kingma2013auto} with Eqn.\  \ref{eqn:forward} to obtain
$\bm{x}_\theta = \frac{1}{\sqrt{\alpha_t}}(\bm{z}_t - \sqrt{1-\alpha_t} \bm{\epsilon}_\theta)$.

After training, we repurpose the denoising neural network into a generative model by starting with Gaussian noise at the maximum noise level $t=1$, i.e., $\bm{z}_1 \sim \mathcal{N}(\bm{0}, \bm{I})$, then iteratively refining the noisy signal, gradually attenuating noise and amplifying signal, by repeatedly computing
\begin{align}
    \hat{\bm{x}}_t &= \frac{1}{\sqrt{\alpha_t}}(\bm{z}_t - \sqrt{1-\alpha_t} \bm{\epsilon}_\theta(\bm{z}_t, t, \bm{c})) \\
    \bm{z}_s &\sim q(\bm{z}_s | \bm{z}_t, \hat{\bm{x}}, \bm{c})~ , ~~~  s<t ~,
\end{align}
for which \citet{ho2020denoising} show that $q(\bm{z}_s | \bm{z}_t, x, \bm{c})$ can be obtained in closed form when $s < t$. To sample with $T$ denoising steps, we typically choose $s$ to be $\frac{T-1}{T}$, then $\frac{T-2}{T}$, and so on, until reaching $s=0$. At the last denoising step, we omit the step that adds noise again, and simply take the final $\hat{\bm{x}}$ to be our sample.

For single-image super-resolution, we used conditional diffusion models.
The data distribution $q(\bm{x},\bm{c})$ is comprised of high-resolution (HR)  images $\bm{x}$ and corresponding low-resolution (LR) images $\bm{c}$.

%% file: relatedwork.tex
\section{Related Work}
\label{section:relatedwork}

Two general approaches to blind super-resolution involve \textit{explicit} \citep{shocher2018zero,liang2021flow,yoo2022rzsr} and \textit{implicit} \citep{patel2021cinc,yan2021fine} degradation modeling. Implicit degradation modeling entails learning the degradation process; however, this requires large datasets to generalize well \cite{liu2021blind}. The best results in the literature employ explicit degradation modeling, where the degradations are directly incorporated as data augmenation during training. \citet{luo2021end,wang2021unsupervised} produce the augmented conditioning images $c$ by applying blur before downsampling the original HR image, and then adding noise and applying JPEG compression to the downsampled result. The Real-ESRGAN model \cite{wang2021real} demonstrates that applying this degradation scheme \textit{more than once} leads to a LR distribution closer to those of images in the wild. These degradation schemes have been crucial for GAN-based methods to achieve state-of-the-art results.

Other methods for super-resolution beyond GANs include diffusion models, and even simpler, non-generative models. The preliminary work of SRCNN \citep{dong2015image} showed the superiority of deep convolutional neural networks over simple bicubic or bilinear upsampling. \citet{dong2016accelerating,shi2016real} improved the efficiency of these results by learning a CNN that itself performs image upsampling. Further architectural and training innovations have since been found to deepen neural networks via residual connections \citep{kim2016accurate,lim2017enhanced,ahn2018fast} and other architectures \citep{8014888,kim2016deeply,8099781,lai2017deep}. Contrastive learning has also been applied to super-resolution \cite{wang2021unsupervised,yin2021conditional}. Attention-based networks have been proposed \citep{8014887,zhang2018image}; however, we still opt to explore a fully convolutional model as it can better generalize to unseen resolutions \citep{whang2022deblurring}. 

Recent work on super-resolution has demonstrated the potential of image-conditional diffusion models \citep{saharia2022image,li2022srdiff}, which were shown to be superior to regression-based models that cannot generate both sharp and diverse samples \citep{ho2022cascaded,saharia2022photorealistic}.
One advantage of diffusion models is their ability to capture the complex statistics of the visual world, as they can  infer structure at scales well beyond those available in LR inputs. This is particularly important at larger magnification factors, where many different HR images may be consistent with a single LR image. By comparison, GAN models often struggle with mode collapse, thereby reducing diversity \citep{thanh2020catastrophic}.

%% file: methodology.tex
\section{Methodology}
\label{section:methodology}

\input{images/degradations.tex}

SR3+ is a self-supervised  diffusion model for blind, single-image super-resolution. 
Its architecture is a convolutional variant of that used in SR3, and hence more flexible with respect to image resolution and aspect ratio. 
During training, it obtains LR-HR image pairs by down-sampling high-resolution images to generate corresponding, low-resolution inputs.
Robustness is achieved through two key augmentations, namely, composite parametric degradations during training \cite{wang2021real,wang2021unsupervised}, and noise conditioning augmentation \citep{ho2022cascaded}, both during training and at test time, as explained below.

\subsection{Architecture}

Following \citet{saharia2022image}, SR3+ uses a UNet architecture, but without the self-attention layers used for SR3.
While self-attention has a positive impact on image quality, it makes generalization to different image resolutions and aspect ratios very difficult \citep{whang2022deblurring}.
We also adopt modifications used by \citet{saharia2022photorealistic} for the Efficient U-Net to improve training speed.
Below we ablate the size of the architecture, demonstrating the performance advantages
of larger models.

\subsection{Higher-order degradations}

Self-supervision for super-resolution entails down-sampling HR images to obtain corresponding LR inputs. Ideally, one combines down-sampling kernels with other degradations that one expects to see in practice.
Otherwise, one can expect a domain shift between training and testing, and hence poor zero-shot generalization to images in the wild.
Arguably, this is a key point of failure of SR3, results of which are evident for ODD test data shown in Figure \ref{fig:representative-samples}.

SR3+ is trained with a data-augmentation pipeline that comprises multiple types of degradation, including image blur, additive noise, JPEG compression and down-sampling.
While the use of multiple parametric deformations in super-resolution training pipelines are common
\citep{zhang2021designing,wang2021unsupervised},
\citet{wang2021real} found that applying 
repeated sequences of deformations, called 
{\em higher-order deformations}, has a substantial impact on OOD generalization.
For simplicity and comparability to Real-ESRGAN, SR3+ uses the same degradation pipeline, but \textit{without} additive noise (see Figure \ref{fig:degradations}).
Empirically, we found in our preliminary experiments that noise conditioning augmentation (explained later) is better than including noise in the degradation pipeline. Training a 400M parameter model on the same dataset as Real-ESRGAN, but with noise in the degradations instead of noise conditioning augmentation, we obtain an FID(10k) score of 42.58 (vs. 36.28, see Table \ref{table:main-results}). For completeness, we now document all the degradation hyperparameters. These should match those used by \citet{wang2021real}.

\textbf{Blur.} Four blur filters are used, i.e., Gaussian, generalized Gaussian, a plateau-based kernel, and a sinc (selected with probabilities 0.63, 0.135, 0.135 and 0.1). With probability $\frac{9}{14}$ the Gaussians are isotropic, and anisotrpic otherwise. The plateau kernel is isotropic with probability $0.8$. When anisotropic, kernels are rotated by a random angle in $(-\pi, \pi]$. 
For isotropic kernels, $\sigma \in [0.2, 3.0]$. For anisotropic kernels, $\sigma_{x},\sigma_{y} \in [0.2, 3.0]$. 
The kernel radius $r$ is random between 3 and 11 pixels (with only with odd value). For the sinc-filter blur, $w_c$ is randomly selected from $[\pi/3, \pi]$ when $r\!<\!6$ and from $[\pi/5, \pi]$ otherwise. For generalized Gaussians, the shape parameter $\beta$ is sampled from $[0.5, 4.0]$; it is sampled from $[1.0, 2.0]$ for the plateau filter. 
The second blur is omitted with probability 0.2; but when used, $\sigma \in [0.2, 1.5]$.

\textbf{Resizing.} Images are resized in one of three (equiprobable) ways, i.e., area resizing, bicubic interpolation, or bilinear interpolation. 
The scale factor is random in $[0.15, 1.5]$ for the first stage resize, and in $[0.3, 1.2]$ for the second. 

\textbf{JPEG compression.} The JPEG quality factor is drawn randomly from $[30, 95]$. In the second stage we also apply a sinc filter (described above), either before or after the JPEG compression (with equal probability).

After two stages of degradations, as illustrated in Fig.\ \ref{fig:degradations}, the image is resized using bicubic interpolation to the desired magnification between the original HR image and the LR degraded image.
SR3+ is trained for $4\times$ magnification.

\subsection{Noise Conditioning Augmentation}

Noise conditioning was first used in  cascaded diffusion models \cite{ho2022cascaded,saharia2022photorealistic}.
It was introduced so that super-resolution models in a cascade can be self-supervised with down-sampling, while at test time it will receive input from the previous model in the cascade. Noise conditioning augmentation provided robustness to the distribution of inputs from the previous stage, even though the stages are trained independently.
While the degradation pipeline should already improve robustness, it is natural to ask whether further robustness can be achieved by also including this technique.

In essence, noise-conditioning augmentation entails addding noise to the up-sampled LR input, but also providing the noise level to the neural denoiser. At training time, for each LR image in a minibatch, it entails
\vspace*{-0.3cm}
\begin{enumerate}[leftmargin=0.5cm]
    \itemsep 0.01cm
    \item Sample $\tau \sim \mathrm{Uniform}(0,\tau_{\mathrm{max}})$.
    \item Add noise to get $\bm{c}_\tau \sim q(\bm{z}_\tau|\bm{c})$, reusing the marginal distribution of the diffusion forward process.
    \item Condition the model on $\bm{c}_\tau$ instead of $\bm{c}$, and we also condition the model on (a positional embedding of) $\tau$.
\end{enumerate}
The model learns to handle input signals at different noise levels $\tau$. In practice, we set $\tau_{\mathrm{max}} = 0.5$; 
beyond this value, the input signal to noise ratio is too low for effective training.

At test time, the noise level
hyper-parameter  in noise-conditioning augmentation, $t_{\mathrm{eval}}$,  provides a trade-off between alignment with the LR input
and hallucination by the generative model.
As $t_{\mathrm{eval}}$ increases, more high-frequency detail is lost, so the model is forced to rely more on its knowledge of natural images than on the conditioning signal per se.
We find that this enables the hallucination of realistic textures and visual detail.

\input{images/model_comparisons.tex}

%% file: images/degradations.tex
\begin{figure*}[t]
\vspace*{-0.1cm}
\begin{center}
\includegraphics[scale=0.25,center]{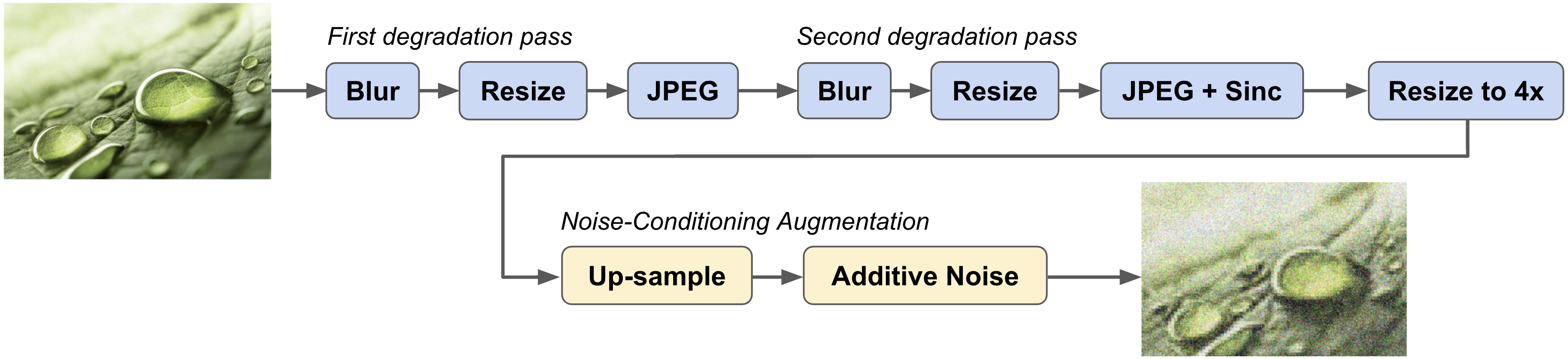}
\vspace*{-0.6cm}
\caption{The SR3+ data  pipeline applies a sequence of degradations to HR training images (like Real-ESRGAN but without additive noise).
To form the conditioning signal for the neural denoiser, we up-sample the LR image and  applied noise conditioning augmentation.
}
\label{fig:degradations}
\end{center}
\vskip -0.3cm
\end{figure*}

%% file: images/model_comparisons.tex
\begin{figure*}[ht!]
\centering
\setlength{\tabcolsep}{2pt}
\resizebox{0.953\textwidth}{!}{%
\begin{tabular}{cccccc}Bicubic & Real-ESRGAN & \textbf{SR3+} (40M) & \textbf{SR3+} (400M) & \textbf{SR3+} (400M,61M)  & Ground Truth\\
\includegraphics[height=1.5in]{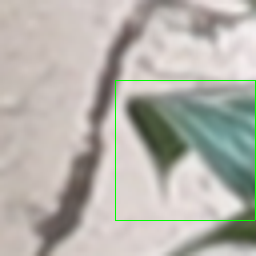} &
\includegraphics[height=1.5in]{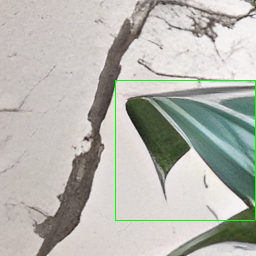} &
\includegraphics[height=1.5in]{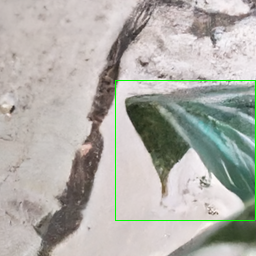} &
\includegraphics[height=1.5in]{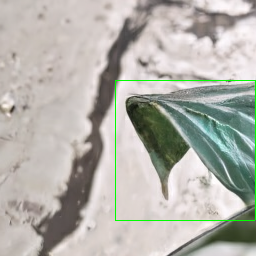} &
\includegraphics[height=1.5in]{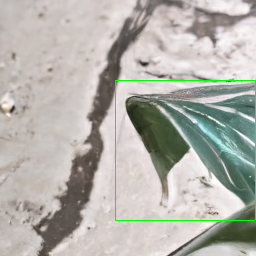} &
\includegraphics[height=1.5in]{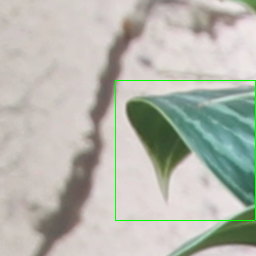} \\
\includegraphics[height=1.5in]{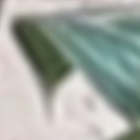} &
\includegraphics[height=1.5in]{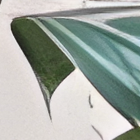} &
\includegraphics[height=1.5in]{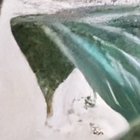} &
\includegraphics[height=1.5in]{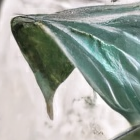} &
\includegraphics[height=1.5in]{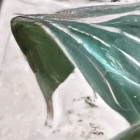} &
\includegraphics[height=1.5in]{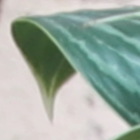} \\
\includegraphics[height=1.5in]{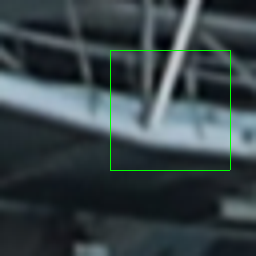} &
\includegraphics[height=1.5in]{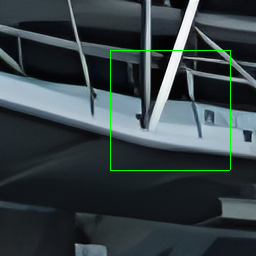} &
\includegraphics[height=1.5in]{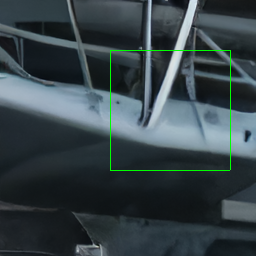} &
\includegraphics[height=1.5in]{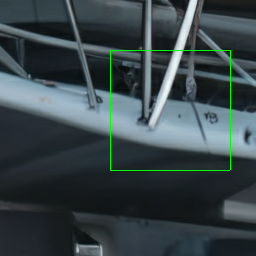} &
\includegraphics[height=1.5in]{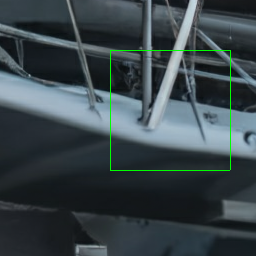} &
\includegraphics[height=1.5in]{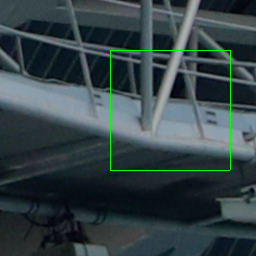} \\
\includegraphics[height=1.5in]{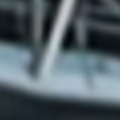} &
\includegraphics[height=1.5in]{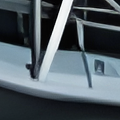} &
\includegraphics[height=1.5in]{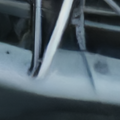} &
\includegraphics[height=1.5in]{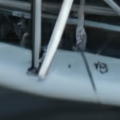} &
\includegraphics[height=1.5in]{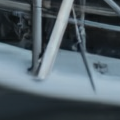} &
\includegraphics[height=1.5in]{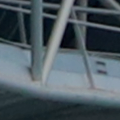} \\
\includegraphics[height=1.5in]{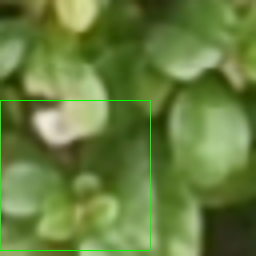} &
\includegraphics[height=1.5in]{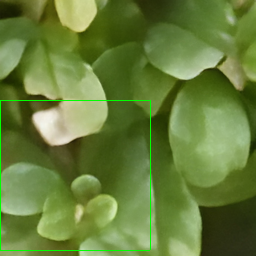} &
\includegraphics[height=1.5in]{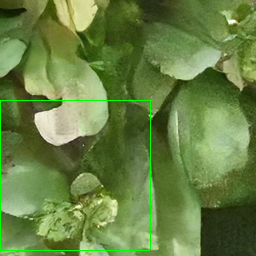} &
\includegraphics[height=1.5in]{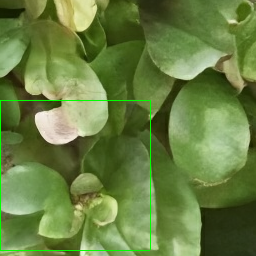} &
\includegraphics[height=1.5in]{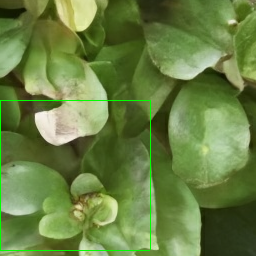} &
\includegraphics[height=1.5in]{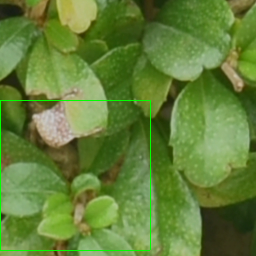} \\
\includegraphics[height=1.5in]{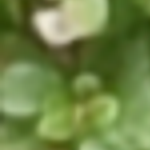} &
\includegraphics[height=1.5in]{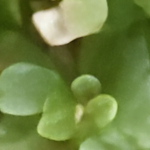} &
\includegraphics[height=1.5in]{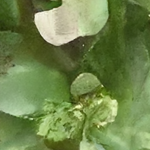} &
\includegraphics[height=1.5in]{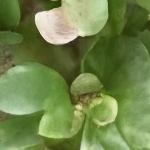} &
\includegraphics[height=1.5in]{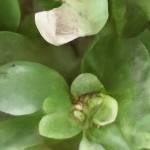} &
\includegraphics[height=1.5in]{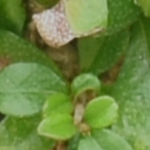} 
\end{tabular}%
} 
\vspace{-0.2cm}
\caption{
Sample comparison between Real-ESRGAN and various SR3+ models (ours). We observe that Real-ESRGAN often suffers from oversmoothing and excessive contrast, while SR3+ is capable of generating high-fidelity, realistic textures.
}
\label{fig:main-results}
\end{figure*}

%% file: experiments.tex
\section{Experiments}
\label{section:experiments}

SR3+  is trained with a combination of degradations and noise-conditioning augmentation on 
multiple  datasets, and applied zero-shot to test data.  We use  ablations to determine the impact of the different forms of augmentation, of model size, and dataset size.
Here, we focus on the blind super-resolution task with a $4\times$ magnification factor. 
For baselines, we use SR3 \cite{saharia2022image} and the previous state-of-the-art in blind super-resolution, i.e., Real-ESRGAN \citep{wang2021real}.

Like SR3, the LR input up-sampled by $4\times$ 
using bicubic interpolation.
The output samples for SR3 and SR3+ are obtained using
DDPM ancestral sampling \citep{ho2020denoising} with 256 denoising steps. For simplicity and to train with continuous timesteps, we use the cosine log-SNR schedule introduced by \citet{ho2022classifier}.

\textbf{Training.}
For fair comparison with Real-ESRGAN, we first train SR3+ on the datasets used to train  Real-ESRGAN \citep{wang2021real}; namely, DF2K+OST \cite{8014884}, a combination of Div2K (800 images), Flick2K (2650 images) and OST300 (300 images).
To explore the impact of scaling, we also train on a large dataset of 61M images, combining a collection of in-house images with DF2K+OST.

\input{tables/comparison.tex}

During training, following Real-ESRGAN, we extract a random $400\!\times\! 400$ crop for each image and then apply 
the degradation pipeline (Fig.\ \ref{fig:degradations}).
The degraded image is then resized to $100\!\times\! 100$ (for $4\times$ magnification).
LR images is then up-sampled using bicubic interpolation to 
$400\!\times\! 400$ from which center crops yield $256\!\times\! 256$ images for training the $64\!\times\! 64 \rightarrow 256\!\times\! 256$ task.  Since the model is convolutional, we can then apply it to arbitrary resolutions and aspect ratios at test time.

For the results below, SR3+ and all ablations are
trained on the same data with the same hyper-parameters. Note that 
SR3+ reduces to SR3 when the degradations and noise-conditioning augmentation are removed.
All models were trained for 1.5M steps, using a batch size of 256 for models trained on DF2K+OST and 512 otherwise. We additionally consider two models sizes, with 40M and 400M weights. The smaller enables direct comparison to Real-ESRGAN, which also has about 40M parameters.
The larger model exposes the impact of model scaling.

\textbf{Testing.} 
For testing, as mentioned above, we focus on zero-shot application to test datasets disjoint from those used for training.
In all experiments and ablations, we use the RealSR \cite{cai2019toward} v3 and DRealSR \cite{wei2020component} datasets for evaluation. 
RealSR has 400 paired low-and-high-resolution images, from which we compute 25 random but aligned 
$64\!\times\! 64$
and
$256\!\times\! 256$
crops per image pair.
This yields a fixed test set of 10,000 image pairs. DRealSR contains more than 10,000 image pairs, so we instead extract $64\!\times\! 64$ and $256\!\times\! 256$ center crops for 10,000 random images.

Model performance is assessed with a combination of PSNR, SSIM \citep{wang2004image} and FID (10k) \citep{heusel2017gans}. 
While reference-based metrics like PSNR and SSIM are useful for small magnification factors, at magnifications of 4x and larger, especially when using a generative model and noise-conditioning augmentation in testing, the posterior distribution is complex, and one expects significant diversity in the output in 
contrast to regression models.

For SR tasks with multi-modal posterios, e.g., at larger magnifications, reference-based 
metrics do not agree well with  human preferences.
While blurry images tend to minimize RMSE from ground truth,  they are scored worse by human observers \citep{chen2018fsrnet,dahl2017pixel,menon2020pulse,saharia2022image}.
In particular PSNR and SSIM tend to over-penalize plausible but infered high-frequency detail that may not agree precisely with ground truth images. We nevertheless consider reconstruction metrics to remain important to evaluate SR models, as they reward alignment and this is a desirable property (especially on regions with less high-frequency details).

In addition to PSNR and SSIM, we also report FID, which on sufficiently larger datasets provides a measure of aggregate statistical similarly with ground truth image data.
This correlates better with human quality assessment.
As generative models are applied to more difficult inputs, or with large amounts of NCA or larger magnifications, we will need to rely more on FID and similar measures.
For in such cases, we will be relying on model inference to capture stats of natural images, and this requires a much larger model, as generative models are hard to learn. So one would expect larger data and larger models would perform better.

\input{images/ablations.tex}

\subsection{Comparison with Real-ESRGAN and SR3}

\input{tables/ablations.tex}

As previously discussed, we compare SR3+ models of different sizes with Real-ESRGAN, the previous state-of-the-art model on blind super-resolution, all trained on the same data. Moreover, in order to attain the best possible results in general, we compare our best SR3+ model trained on said data with an identical one that was instead trained on the much larger 61M-image dataset (and with twice the batch size). For evaluation, we perform a grid sweep over $t_{\mathrm{eval}}$ from 0 to 0.4, with increments of 0.05, and report results with $t_{\mathrm{eval}}=0.1$, which we consistently find to be the best value. We provide side-by-side comparisons in Figure \ref{fig:main-results}, and show quantitative results in Table \ref{table:main-results}.

We find that, with a 40M-parameter network, SR3+ achieves competitive FID scores with Real-ESRGAN, achieving better scores on RealSR but slightly worse on DRealSR. Qualitatively, it creates more realistic textures without significant oversmoothing or saturation, but it does worse for certain kinds of images where we care about accurate high-frequency detail, such as images with text. The results and realism of the images improve significantly with a 400M-parameter SR3+ model, outperforming Real-ESRGAN on FID scores when trained on the same dataset, and this gap is furthered widened simply by training on the much larger dataset. In the latter case, some of the failure modes of the earlier models (e.g., the text case) are also alleviated, and rougher textures are more coherent within the images. We provide additional samples in the Supplementary Material.

SR3+ does not outperform on reference-based metrics (PSNR, SSIM) are slightly worse, but this expected from strong generative models with either larger magnification factors or larger noise-conditioning augmentation (where the generative model is forced to infer more details). This is also shown by prior work \citep{chen2018fsrnet,dahl2017pixel,menon2020pulse,saharia2022image}. We verify this empirically in the samples shown in Figure \ref{fig:ablations} and Table \ref{table:ablations}, where,
notably, SR3 attains better PSNR and SSIM scores, but the model produces blurry results in the blind task.
In the 4x magnification task starting from 64x64, $p(\bm{x}|\bm{c})$ can be very multimodal (especially on high-frequency details), and these metrics overpenalize plausible but hallucinated high-frequency details.

\subsection{Ablation studies}

We now empirically demonstrate the importance of our main contributions, which we recall are (1) the higher-order degradation scheme and (2) noise conditioning augmentation. We conduct an ablation study using our strongest model, i.e., the 400M-parameter SR3+ model trained on the 61M-image dataset, as worse models break more dramatically upon removing said components. We train similar models as our strongest SR3+ model: one without noise conditioning augmentation, one without the higher-order degradations, and one with neither (which is equivalent to an SR3 model, though using the UNetv3 architecture \citep{saharia2022photorealistic} and in a larger dataset than in the original work). We then compare FID, PSNR and SSIM on the blind SR task, as before. Whenever using noise-conditioning augmentation, we set $t_{\mathrm{eval}}=0.1$. Results are included in Table \ref{table:ablations} and a sample comparison in Figure \ref{fig:ablations}.



Our results show that FID scores increase significantly upon the removal of either of our main contributions (by over 10 points in all cases). And, upon removing both, FID scores are much worse, as this metric punishes the consistent blurriness of SR3 when applied in the wild to out-of-distribution images. We also observe that, specifically without the higher-order degradations, we also observe some bluriness and a slight improvement across reconstruction metrics. With the SR3 model, which qualitatively appears to suffer most from blur in generations, both PSNR and SSIM improve significantly, and, interestingly enough, sufficiently to outperform Real-ESRGAN in both metrics and both evaluation datasets.


\subsection{Noise conditioning augmentation at test time}
Recall that, due to the use of noise conditioning augmentation, we introduce a degree of freedom $t_{\mathrm{eval}}$ at sampling time that we are free to play with. Intuitively, it would seem that using $t_{\mathrm{eval}} = 0$ would be most appropriate, as adding noise removes some information from the conditioning low-resolution input. Empirically, however, we find that using a nonzero $t_{\mathrm{eval}}$ can often lead to better results; especially on images where highly detailed textures are desirable. To demonstrate this, we present a comparison of FID scores across different values of $t_{\mathrm{eval}}$ in Figure \ref{fig:t-eval-sweep}, for our two 400M-parameter SR3+ models (recall, one trained on DF2K+OST and one on the 61M-image dataset). We additionally include samples from the SR3+ model trained on the 61M-image dataset in Figure 
\ref{fig:t-eval-samples}.

\input{images/t_eval.tex}
\input{images/FID_vs_teval.tex}

For both models and both evaluation datasets, we find that FID scores can visibly drop when using noise conditioning augmentation at test time, with the best value often about $t_{\mathrm{eval}} = 0.1$. With the model trained on the 61M-image dataset, we curiously find that more aggressive noise conditioning augmentation can be used at test time while still attaining better FID scores than with $t_{\mathrm{eval}} = 0$. In our samples, we show that the effect of using small amounts of test-time noise conditioning augmentation has a subtle but beneficial effect: higher-quality textures appear and there is less bluriness than without any noise, and alignment to the conditioning image remains good or can even improve (e.g., the flower pot seemed to shift up without noise). As we increase $t_{\mathrm{eval}}$, however, we begin to see initially small but increasingly more apparent misalignment to the conditioning image, as more high-frequency information is destroyed with increasing amounts of noise applied to the conditioning signal. This forces the model to rely on its own knowledge to hallucinate such details and textures, which can be beneficial in most cases (but less so with, e.g., text).

%% file: tables/comparison.tex
\begin{table*}[t]
\begin{center}
\begin{tabular}{ |c|c|c|c|c|c|c| } 
 \hline
 \multirow{2}{*}{SR Model (Parameter Count, Dataset)} & \multicolumn{2}{c|}{FID(10k)	$\downarrow$} & \multicolumn{2}{c|}{PSNR	$\uparrow$} & \multicolumn{2}{c|}{SSIM	$\uparrow$} \\
 \cline{2-7}
 & RealSR & DRealSR & RealSR & DRealSR & RealSR & DRealSR \\
 \hline
 Real-ESRGAN & 34.21 & 37.22 & \textbf{25.14} & \textbf{25.85} & \textbf{0.7279} & \textbf{0.7808} \\ 
 SR3+ (40M, DF2K + OST) & \underline{31.97} & 40.26 & 24.84 & 25.18 & 0.6827 & 0.7201 \\
 SR3+ (400M, DF2K + OST) & \underline{27.34} & \underline{36.28} & 23.84 & 24.36 & 0.662 & 0.719 \\
 SR3+ (400M, 61M Dataset) & \textbf{24.32} & \textbf{32.37} & 24.89 & 25.74 & 0.6922 & 0.7547 \\
 \hline
\end{tabular}
\end{center}
\vspace*{-0.25cm}
\caption{Quantitative comparison between Real-ESRGAN and SR3+ (ours). We achieve similar FID scores with a 40M parameter model, and find significant improvement upon increasing model and dataset sizes.}
\label{table:main-results}
\end{table*}

%% file: images/ablations.tex


\begin{figure*}[ht]
\centering
\setlength{\tabcolsep}{2pt}
\resizebox{\textwidth}{!}{
\small
\begin{tabular}{ccccc}
Input & \textbf{SR3+} & No noise cond.\ aug.\ & No degradations & SR3 (ablate both)\\
\includegraphics[height=1in]{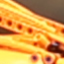} &
\includegraphics[height=1in]{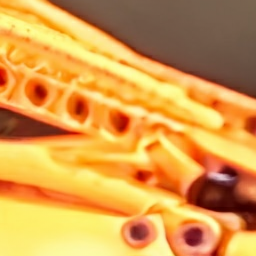} &
\includegraphics[height=1in]{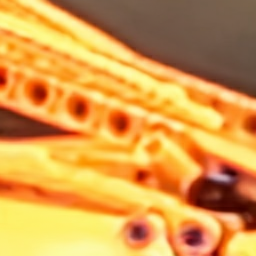} &
\includegraphics[height=1in]{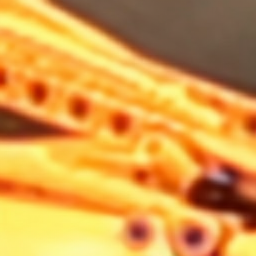} &
\includegraphics[height=1in]{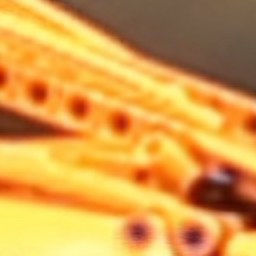}\\
\vspace*{-0.35cm} \\
\includegraphics[height=1in]{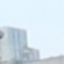} &
\includegraphics[height=1in]{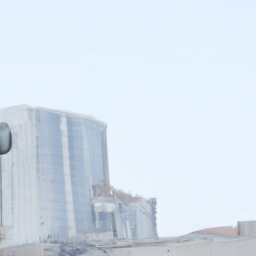} &
\includegraphics[height=1in]{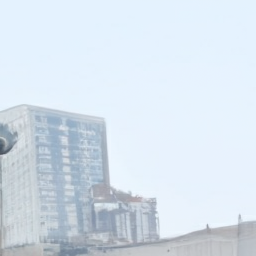} &
\includegraphics[height=1in]{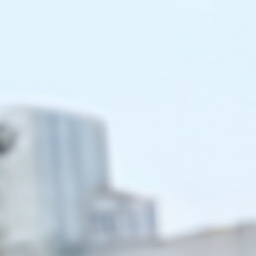} &
\includegraphics[height=1in]{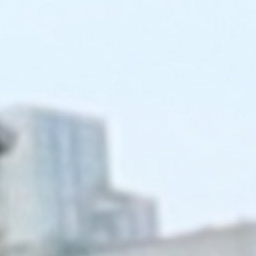} \\
\vspace*{-0.35cm} \\
\includegraphics[height=1in]{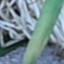} &
\includegraphics[height=1in]{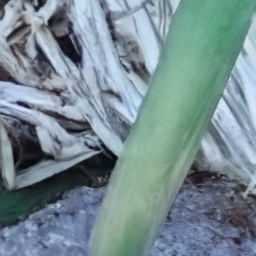} &
\includegraphics[height=1in]{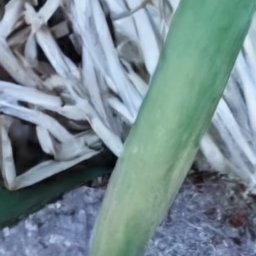} &
\includegraphics[height=1in]{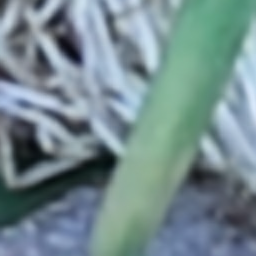} &
\includegraphics[height=1in]{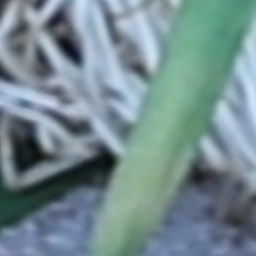} 
\end{tabular}
}
\vspace*{-0.3cm}
\caption{Ablation samples ($t_{eval}\!=\! 0.1$), illustrating the importance of higher-order degradations and noise conditioning augmentation. 
}
\label{fig:ablations}
\end{figure*}

%% file: tables/ablations.tex
\begin{table*}[t]
\begin{center}
\begin{tabular}{ |c|c|c|c|c|c|c| }
 \hline
 \multirow{2}{*}{SR Model (400M parameters, 61M Dataset)} & \multicolumn{2}{c|}{FID(10k) $\downarrow$} & \multicolumn{2}{c|}{PSNR$\uparrow$} & \multicolumn{2}{c|}{SSIM$\uparrow$} \\ 
 \cline{2-7}
 & RealSR & DRealSR & RealSR & DRealSR & RealSR & DRealSR \\
 \hline
 SR3+ & \textbf{24.32} & \textbf{32.37} & 24.89 & 25.74 & 0.6922 & 0.7547 \\ \hline
  \shortstack{SR3+ (no noise cond. aug.)} & 34.20 & 49.93 & 22.34 & 22.28 & 0.6469 & 0.6994 \\ \hline
 \shortstack{SR3+ (no degradations)} & 36.93 & 44.18 & \underline{25.00} & \underline{26.22} & 0.6824 & \underline{0.7687} \\
 \hline
 \shortstack{SR3 (i.e., ablating both)} & 85.77 & 93.05 & \textbf{27.89} & \textbf{28.25} & \textbf{0.784} & \textbf{0.83} \\ \hline
\end{tabular}
\end{center}
\vspace*{-0.2cm}
\caption{Ablation study over SR3+ on the RealSR and DRealSR test sets. Note that ablating both components yields the SR3 model.}
\label{table:ablations}
\end{table*}

%% file: images/t_eval.tex
\begin{figure*}[ht]
\centering
\setlength{\tabcolsep}{2pt}
\resizebox{\textwidth}{!}{%
\begin{tabular}{ccccccc} \\
Input & $t_{eval}=0.0$ & $t_{eval}=0.1$ & $t_{eval}=0.2$ & $t_{eval}=0.3$ & $t_{eval}=0.4$ & Ground truth \\
\includegraphics[height=1in]{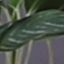} &
\includegraphics[height=1in]{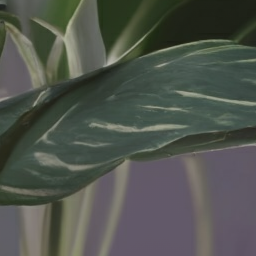} &
\includegraphics[height=1in]{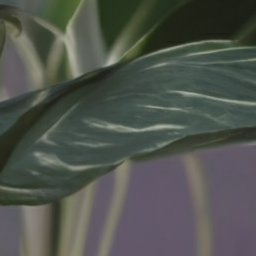} &
\includegraphics[height=1in]{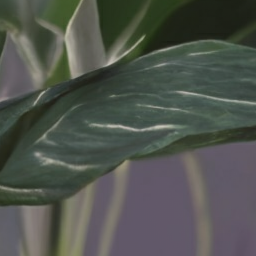} &
\includegraphics[height=1in]{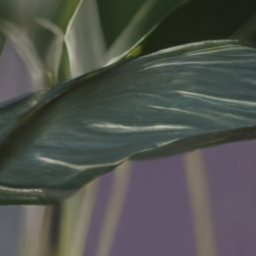} &
\includegraphics[height=1in]{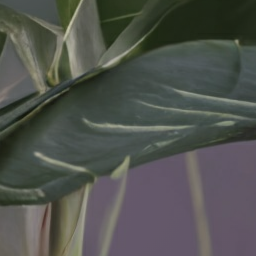} &
\includegraphics[height=1in]{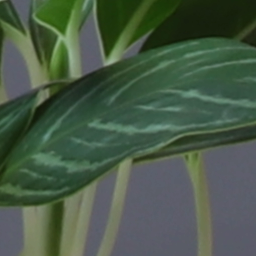} \\
\includegraphics[height=1in]{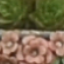} &
\includegraphics[height=1in]{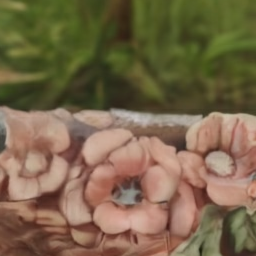} &
\includegraphics[height=1in]{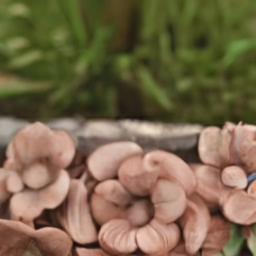} &
\includegraphics[height=1in]{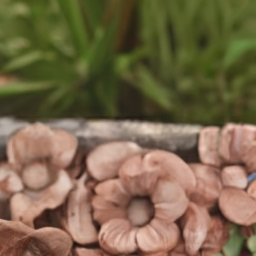} &
\includegraphics[height=1in]{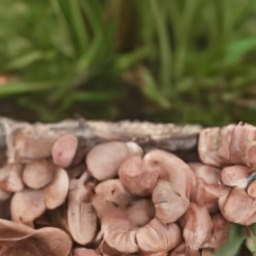} &
\includegraphics[height=1in]{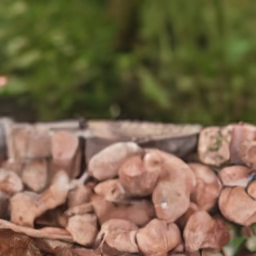} &
\includegraphics[height=1in]{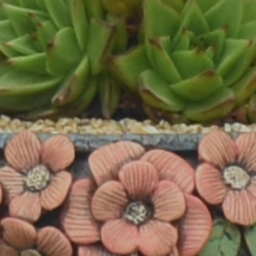} \\
\includegraphics[height=1in]{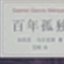} &
\includegraphics[height=1in]{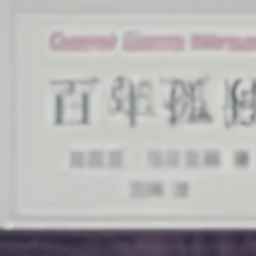} &
\includegraphics[height=1in]{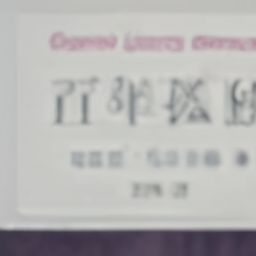} &
\includegraphics[height=1in]{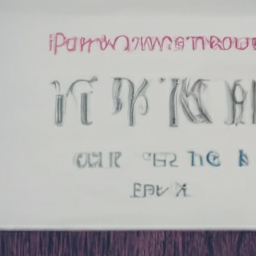} &
\includegraphics[height=1in]{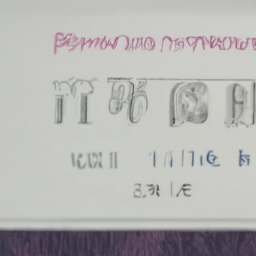} &
\includegraphics[height=1in]{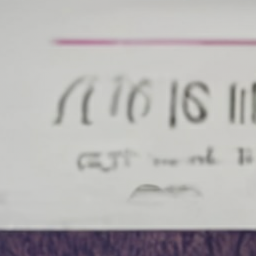} &
\includegraphics[height=1in]{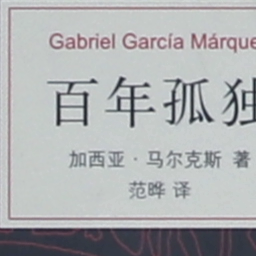} \\
\end{tabular}%
}
\vspace{-0.5cm}
\caption{Samples from SR3+ (400M weights, 61M dataset) using different amounts of test-time noise conditioning augmentation, $t_{eval}$.}
\label{fig:t-eval-samples}
\end{figure*}

%% file: images/FID_vs_teval.tex
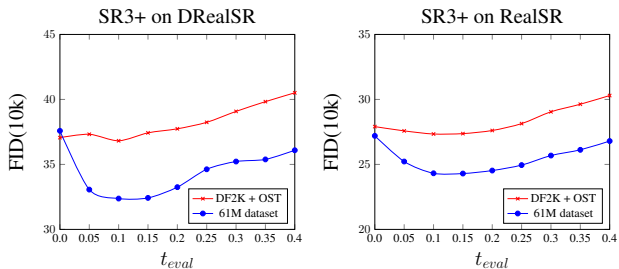
\begin{figure}[b]
\resizebox{1.025\columnwidth}{!}{
    \input{plot_teval_fid_400M_bigdataset.tex}
    \input{plot_teval_fid_400M_df2kost.tex}
}  
\vspace*{-0.8cm}
\caption{FID score comparisons for different ammounts of test-time noise conditioning augmentation. We include results with two 400M-parameter SR3+ models, one trained on the DF2K+OST dataset, and another trained on the much larger 61M-image dataset.}
\label{fig:t-eval-sweep}
\end{figure}

%% file: plot_teval_fid_400M_bigdataset.tex
\resizebox{0.479\columnwidth}{!}{%
\begin{tikzpicture}
\begin{axis}[
    title=\LARGE{SR3+ on DRealSR},
    xlabel=\LARGE{$t_{eval}$},
    ylabel=\LARGE{FID(10k)},
    xmin=0.0, xmax=0.4,
    ymin=30, ymax=45,
    legend pos=south east,
    xtick={0.0,0.05,0.1,0.15,0.2,0.25,0.3,0.35,0.4,0.45,0.5},
    xticklabels={0.0,0.05,0.1,0.15,0.2,0.25,0.3,0.35,0.4,0.45,0.5},
    ytick={30,35,40,45}
]
\addplot[smooth,mark=x,red] plot coordinates {
    (0.0,37.06)
    (0.05,37.32)
    (0.10,36.82)
    (0.15,37.43)
    (0.20,37.74)
    (0.25,38.24)
    (0.30,39.08)
    (0.35,39.83)
    (0.40,40.51)
};
\addlegendentry{DF2K + OST}

\addplot[smooth,mark=*,blue] plot coordinates {
    (0.0,37.584)
    (0.05,33.063)
    (0.10,32.374)
    (0.15,32.424)
    (0.20,33.250)
    (0.25,34.625)
    (0.30,35.222)
    (0.35,35.385)
    (0.40,36.089)
};
\addlegendentry{61M dataset}
\end{axis}
\end{tikzpicture}%
}

%% file: plot_teval_fid_400M_df2kost.tex
\resizebox{0.479\columnwidth}{!}{%
\begin{tikzpicture}
\begin{axis}[
    title=\LARGE{SR3+ on RealSR},
    xlabel=\LARGE{}{$t_{eval}$},
    ylabel=\LARGE{FID(10k)},
    xmin=0.0, xmax=0.4,
    ymin=20, ymax=35,
    legend pos=south east,
    xtick={0.0,0.05,0.1,0.15,0.2,0.25,0.3,0.35,0.4,0.45,0.5},
    xticklabels={0.0,0.05,0.1,0.15,0.2,0.25,0.3,0.35,0.4,0.45,0.5},
    ytick={20,25,30,35,40,...,100}
]

\addplot[smooth,color=red,mark=x]
    plot coordinates {
        (0.0,27.91)
        (0.05,27.58)
        (0.10,27.34)
        (0.15,27.37)
        (0.20,27.61)
        (0.25,28.14)
        (0.30,29.05)
        (0.35,29.63)
        (0.40,30.30)
    };
\addlegendentry{DF2K + OST}

\addplot[smooth,color=blue,mark=*]
    plot coordinates {
        (0.0,27.199)
        (0.05,25.223)
        (0.10,24.317)
        (0.15,24.298)
        (0.20,24.529)
        (0.25,24.948)
        (0.30,25.682)
        (0.35,26.123)
        (0.40,26.798)
    };
\addlegendentry{61M dataset}
\end{axis}
\end{tikzpicture}%
}

%% file: conclusion.tex
\section{Conclusion}
\label{section:conclusion}
  In this work, we propose SR3+, a diffusion model for blind super-resolution. By combining two recent techniques for image enhancement, a higher order degradation scheme and noise conditioning augmentation,
  SR3+ achieves state-of-the-art FID scores across test datasets for blind super-resolution. We further improve quantitative and qualitative results significantly just by training on a much larger dataset. Unlike prior work, SR3+ is both robust to out-of-distribution inputs, and can generate realistic textures in a controllable manner, as test-time noise conditioning augmentation can force the model to rely on more of its own knowledge to infer high-frequency details. SR3+ excels at natural images, and with enough data, it performs reasonably well on other images such as those with text. We are most excited about SR3+ improving diffusion model quality and robustness more broadly, especially those relying on cascading \citep{ho2022cascaded}, e.g., text-to-image models.
  
  SR3+ nevertheless has some limitations. 
  When using noise conditioning augmentation, some failure modes can be observed such as gibberish text, and more training steps might needed for convergence as the task becomes more challenging than with conditioning signals that are always clean. 
  We believe that models with larger capacity (i.e., parameter count), as well as improvements on neural architectures, could address these issues in future work.

%% file: appendix.tex
\section{Additional sample comparisons between SR3+ and other models}
We provide more side-by-side comparisons of our best SR3+ model with SR3 and Real-ESRGAN. We display a single zoomed-in crop for each image to emphasize that SR3+ outperforms prior work.
\label{appendix:samples}

\begin{figure}[h!]
\vspace{-0.5cm}
\centering
\setlength{\tabcolsep}{2pt}
\resizebox{0.97\textwidth}{!}{%
\begin{tabular}{cccccc} \\
Input & SR3 & Real-ESRGAN & \textbf{SR3+ (Ours)} & Ground Truth\\

\includegraphics[height=1in]{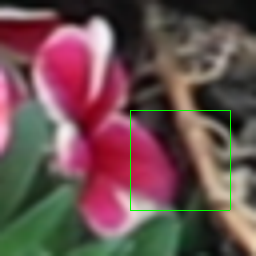} &
\includegraphics[height=1in]{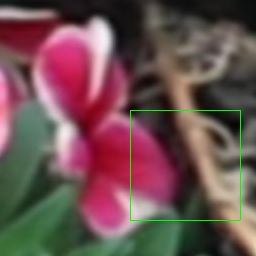} &
\includegraphics[height=1in]{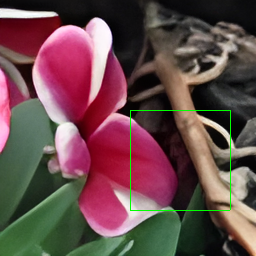} &
\includegraphics[height=1in]{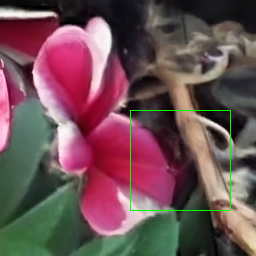} &
\includegraphics[height=1in]{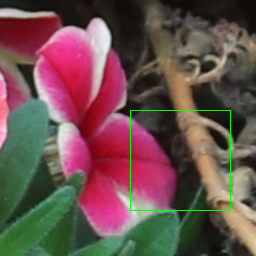} \\
\includegraphics[height=1in]{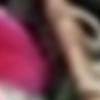} &
\includegraphics[height=1in]{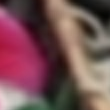} &
\includegraphics[height=1in]{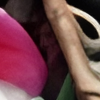} &
\includegraphics[height=1in]{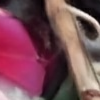} &
\includegraphics[height=1in]{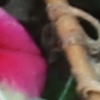} \\

\includegraphics[height=1in]{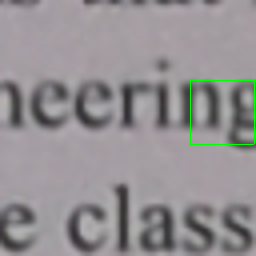} &
\includegraphics[height=1in]{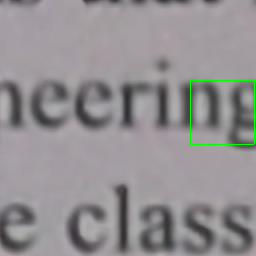} &
\includegraphics[height=1in]{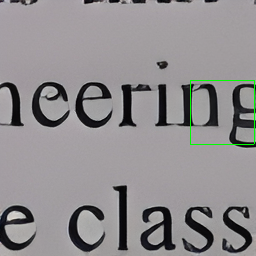} &
\includegraphics[height=1in]{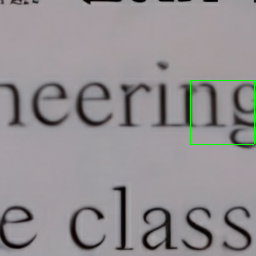} &
\includegraphics[height=1in]{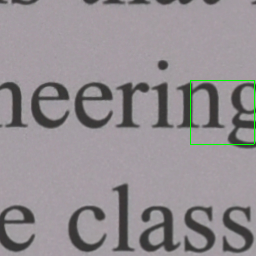} \\
\includegraphics[height=1in]{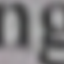} &
\includegraphics[height=1in]{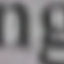} &
\includegraphics[height=1in]{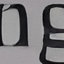} &
\includegraphics[height=1in]{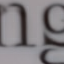} &
\includegraphics[height=1in]{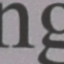} \\

\includegraphics[height=1in]{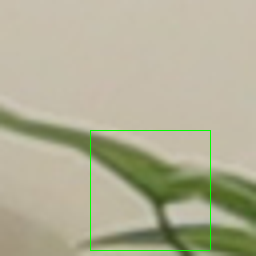} &
\includegraphics[height=1in]{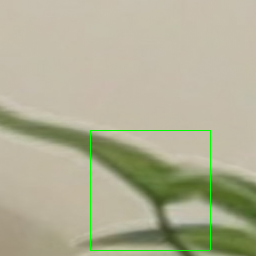} &
\includegraphics[height=1in]{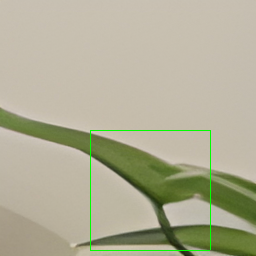} &
\includegraphics[height=1in]{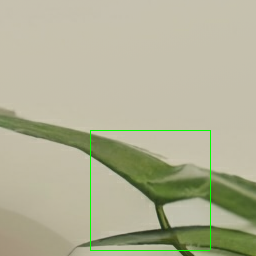} &
\includegraphics[height=1in]{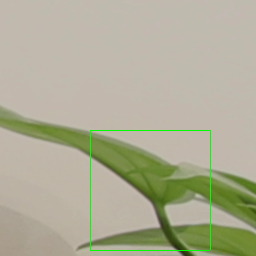} \\
\includegraphics[height=1in]{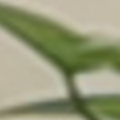} &
\includegraphics[height=1in]{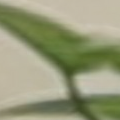} &
\includegraphics[height=1in]{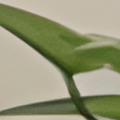} &
\includegraphics[height=1in]{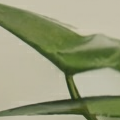} &
\includegraphics[height=1in]{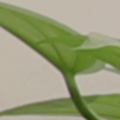} \\
\end{tabular}%
}
\end{figure}

\begin{figure}[ht]
\centering
\setlength{\tabcolsep}{2pt}
\resizebox{\textwidth}{!}{%
\begin{tabular}{cccccc} \\
Input & SR3 & Real-ESRGAN & \textbf{SR3+ (Ours)} & Ground Truth\\

\includegraphics[height=1in]{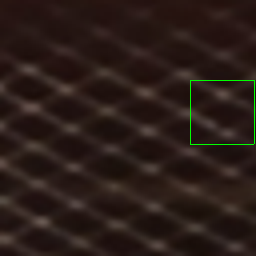} &
\includegraphics[height=1in]{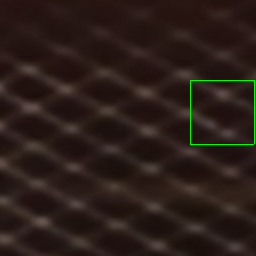} &
\includegraphics[height=1in]{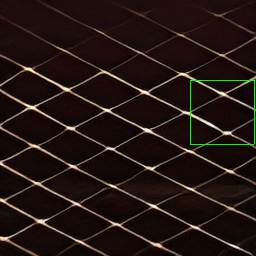} &
\includegraphics[height=1in]{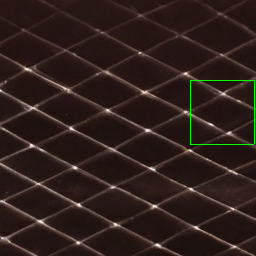} &
\includegraphics[height=1in]{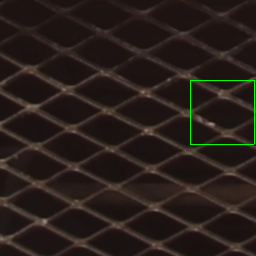} \\
\includegraphics[height=1in]{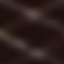} &
\includegraphics[height=1in]{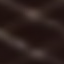} &
\includegraphics[height=1in]{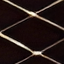} &
\includegraphics[height=1in]{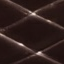} &
\includegraphics[height=1in]{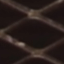} \\

\includegraphics[height=1in]{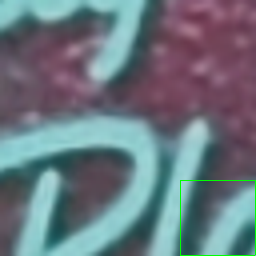} &
\includegraphics[height=1in]{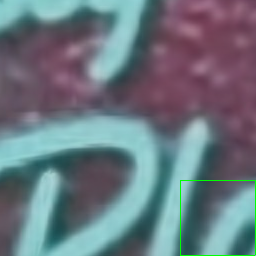} &
\includegraphics[height=1in]{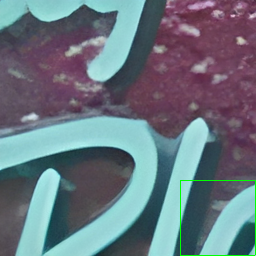} &
\includegraphics[height=1in]{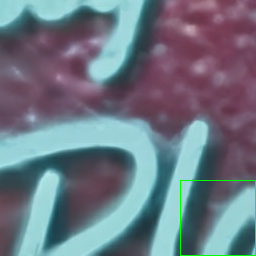} &
\includegraphics[height=1in]{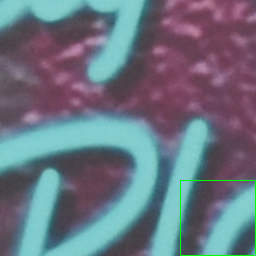} \\
\includegraphics[height=1in]{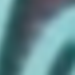} &
\includegraphics[height=1in]{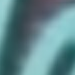} &
\includegraphics[height=1in]{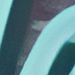} &
\includegraphics[height=1in]{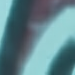} &
\includegraphics[height=1in]{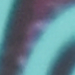} \\

\includegraphics[height=1in]{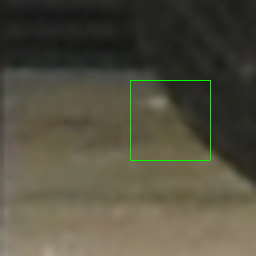} &
\includegraphics[height=1in]{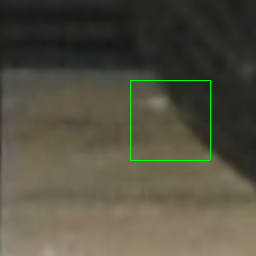} &
\includegraphics[height=1in]{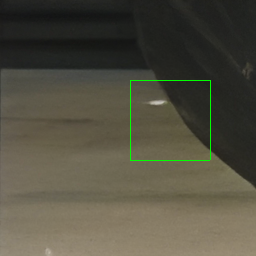} &
\includegraphics[height=1in]{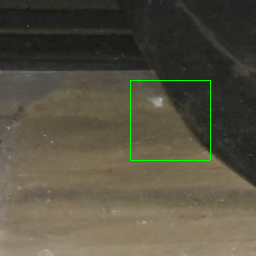} &
\includegraphics[height=1in]{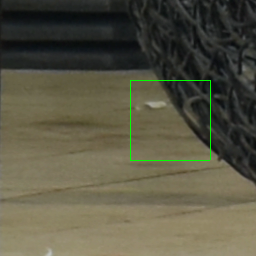} \\
\includegraphics[height=1in]{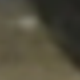} &
\includegraphics[height=1in]{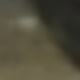} &
\includegraphics[height=1in]{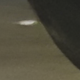} &
\includegraphics[height=1in]{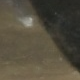} &
\includegraphics[height=1in]{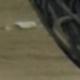} \\
\end{tabular}%
}
\end{figure}

\begin{figure}[ht]
\centering
\setlength{\tabcolsep}{2pt}
\resizebox{\textwidth}{!}{%
\begin{tabular}{cccccc} \\
Input & SR3 & Real-ESRGAN & \textbf{SR3+ (Ours)} & Ground Truth\\

\includegraphics[height=1in]{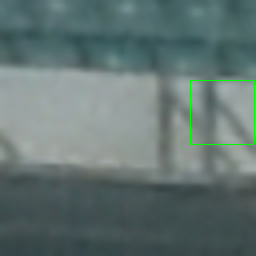} &
\includegraphics[height=1in]{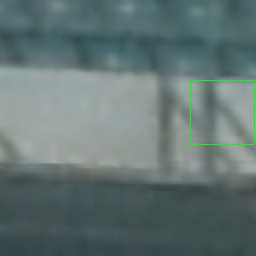} &
\includegraphics[height=1in]{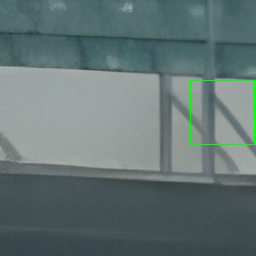} &
\includegraphics[height=1in]{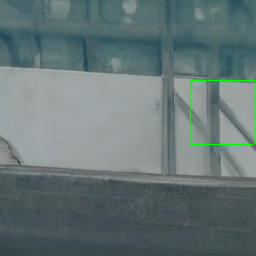} &
\includegraphics[height=1in]{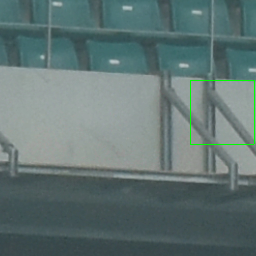} \\
\includegraphics[height=1in]{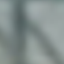} &
\includegraphics[height=1in]{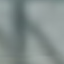} &
\includegraphics[height=1in]{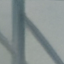} &
\includegraphics[height=1in]{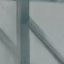} &
\includegraphics[height=1in]{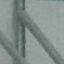} \\

\includegraphics[height=1in]{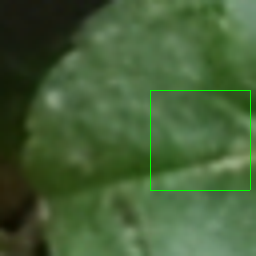} &
\includegraphics[height=1in]{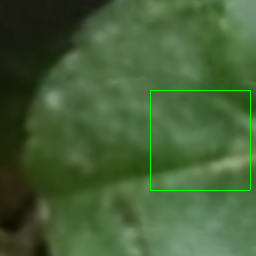} &
\includegraphics[height=1in]{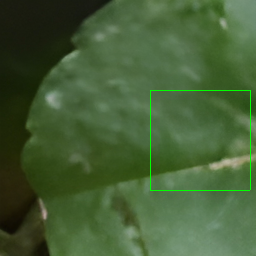} &
\includegraphics[height=1in]{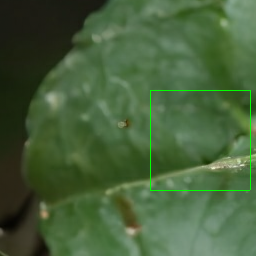} &
\includegraphics[height=1in]{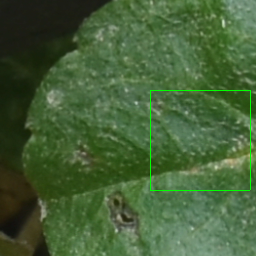} \\
\includegraphics[height=1in]{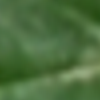} &
\includegraphics[height=1in]{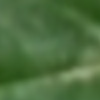} &
\includegraphics[height=1in]{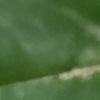} &
\includegraphics[height=1in]{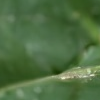} &
\includegraphics[height=1in]{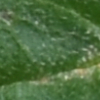} \\

\includegraphics[height=1in]{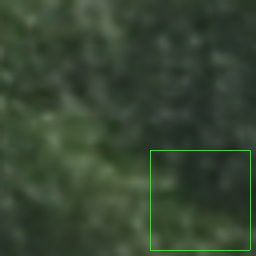} &
\includegraphics[height=1in]{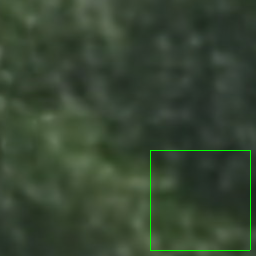} &
\includegraphics[height=1in]{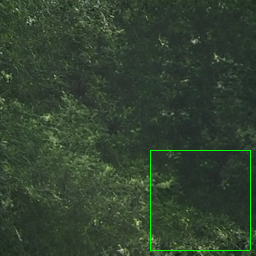} &
\includegraphics[height=1in]{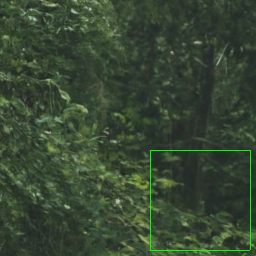} &
\includegraphics[height=1in]{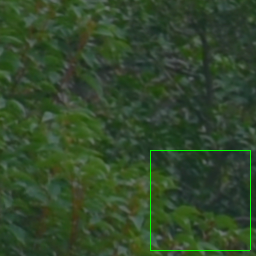} \\
\includegraphics[height=1in]{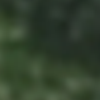} &
\includegraphics[height=1in]{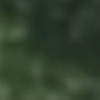} &
\includegraphics[height=1in]{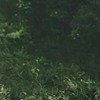} &
\includegraphics[height=1in]{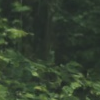} &
\includegraphics[height=1in]{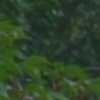} \\
\end{tabular}%
}
\end{figure}

\begin{figure}[ht]
\centering
\setlength{\tabcolsep}{2pt}
\resizebox{\textwidth}{!}{%
\begin{tabular}{cccccc} \\
Input & SR3 & Real-ESRGAN & \textbf{SR3+ (Ours)} & Ground Truth\\

\includegraphics[height=1in]{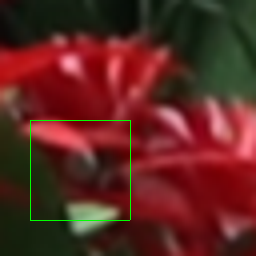} &
\includegraphics[height=1in]{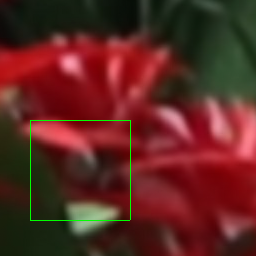} &
\includegraphics[height=1in]{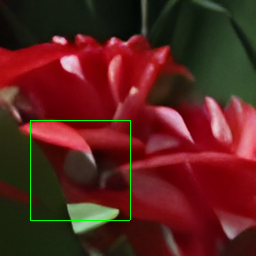} &
\includegraphics[height=1in]{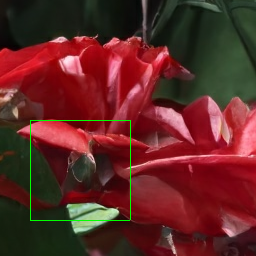} &
\includegraphics[height=1in]{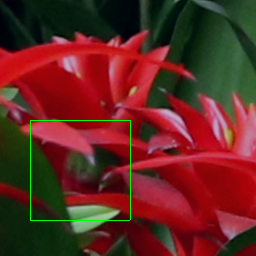} \\
\includegraphics[height=1in]{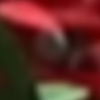} &
\includegraphics[height=1in]{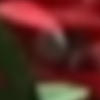} &
\includegraphics[height=1in]{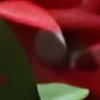} &
\includegraphics[height=1in]{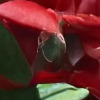} &
\includegraphics[height=1in]{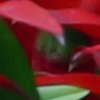} \\

\includegraphics[height=1in]{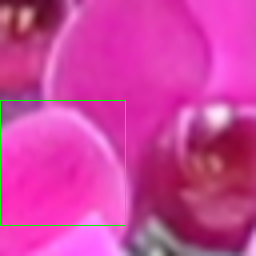} &
\includegraphics[height=1in]{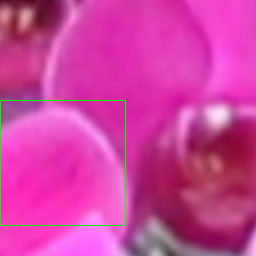} &
\includegraphics[height=1in]{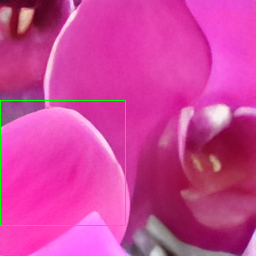} &
\includegraphics[height=1in]{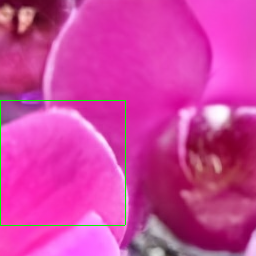} &
\includegraphics[height=1in]{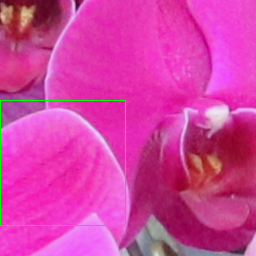} \\
\includegraphics[height=1in]{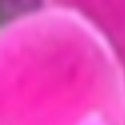} &
\includegraphics[height=1in]{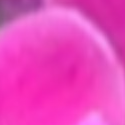} &
\includegraphics[height=1in]{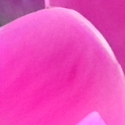} &
\includegraphics[height=1in]{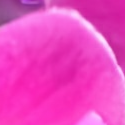} &
\includegraphics[height=1in]{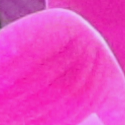} \\

\includegraphics[height=1in]{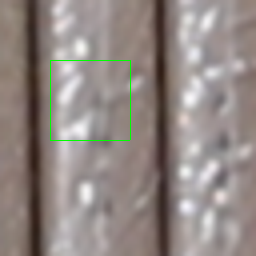} &
\includegraphics[height=1in]{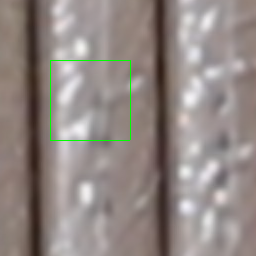} &
\includegraphics[height=1in]{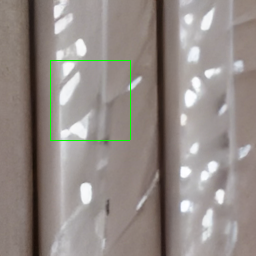} &
\includegraphics[height=1in]{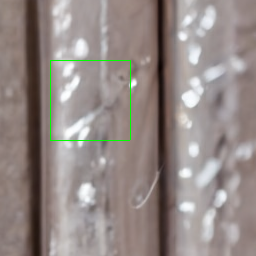} &
\includegraphics[height=1in]{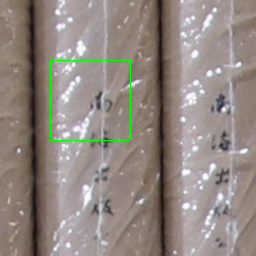} \\
\includegraphics[height=1in]{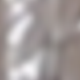} &
\includegraphics[height=1in]{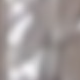} &
\includegraphics[height=1in]{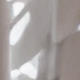} &
\includegraphics[height=1in]{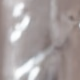} &
\includegraphics[height=1in]{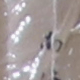} \\
\end{tabular}%
}
\end{figure}

\begin{figure}[ht]
\centering
\setlength{\tabcolsep}{2pt}
\resizebox{\textwidth}{!}{%
\begin{tabular}{cccccc} \\
Input & SR3 & Real-ESRGAN & \textbf{SR3+ (Ours)} & Ground Truth\\

\includegraphics[height=1in]{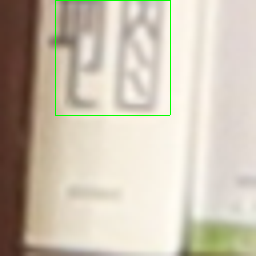} &
\includegraphics[height=1in]{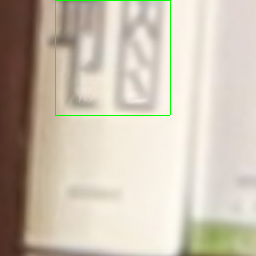} &
\includegraphics[height=1in]{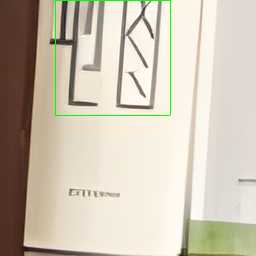} &
\includegraphics[height=1in]{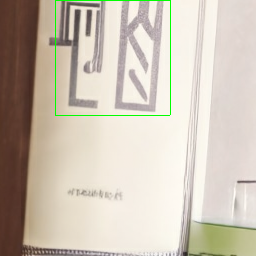} &
\includegraphics[height=1in]{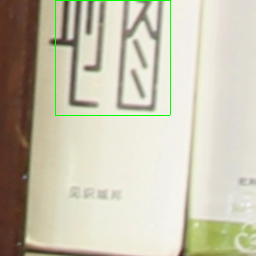} \\
\includegraphics[height=1in]{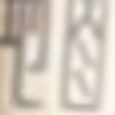} &
\includegraphics[height=1in]{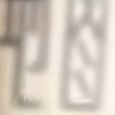} &
\includegraphics[height=1in]{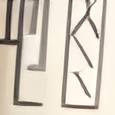} &
\includegraphics[height=1in]{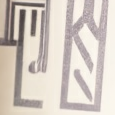} &
\includegraphics[height=1in]{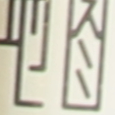} \\

\includegraphics[height=1in]{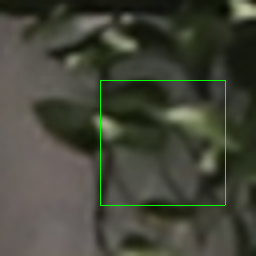} &
\includegraphics[height=1in]{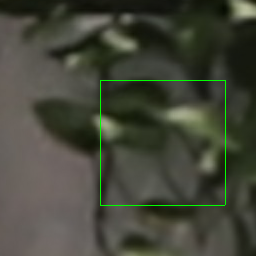} &
\includegraphics[height=1in]{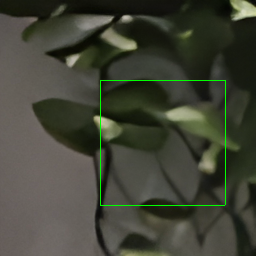} &
\includegraphics[height=1in]{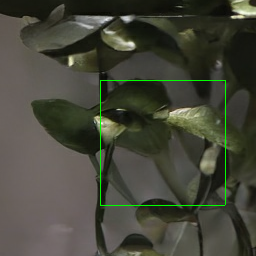} &
\includegraphics[height=1in]{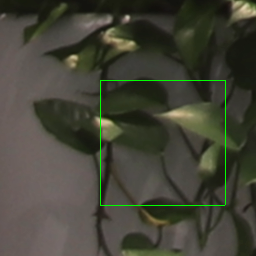} \\
\includegraphics[height=1in]{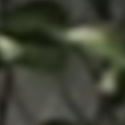} &
\includegraphics[height=1in]{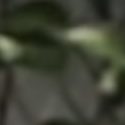} &
\includegraphics[height=1in]{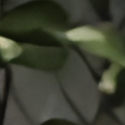} &
\includegraphics[height=1in]{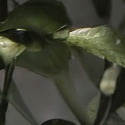} &
\includegraphics[height=1in]{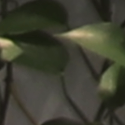} \\

\includegraphics[height=1in]{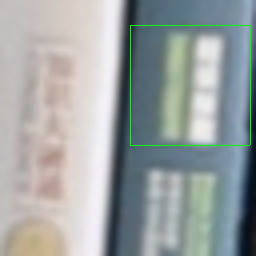} &
\includegraphics[height=1in]{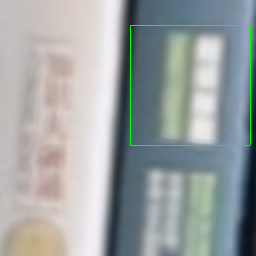} &
\includegraphics[height=1in]{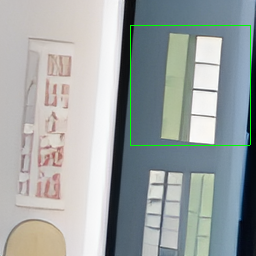} &
\includegraphics[height=1in]{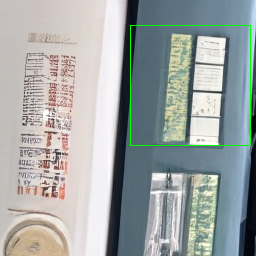} &
\includegraphics[height=1in]{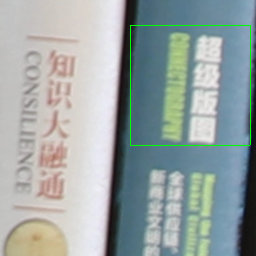} \\
\includegraphics[height=1in]{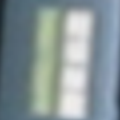} &
\includegraphics[height=1in]{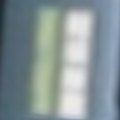} &
\includegraphics[height=1in]{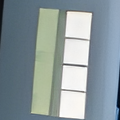} &
\includegraphics[height=1in]{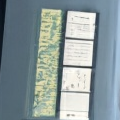} &
\includegraphics[height=1in]{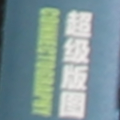} \\
\end{tabular}%
}
\end{figure}